\documentclass{article}
\usepackage[final]{colm2026_conference}

\usepackage{microtype}
\usepackage{hyperref}
\usepackage{url}
\usepackage{booktabs}
\usepackage{lineno}

\definecolor{darkblue}{rgb}{0, 0, 0.5}
\hypersetup{colorlinks=true, citecolor=darkblue, linkcolor=darkblue, urlcolor=darkblue}

\usepackage{amsfonts}
\usepackage{nicefrac}
\usepackage{xcolor}
\usepackage{tabularx}
\usepackage{tcolorbox}
\tcbuselibrary{skins,breakable}
\usepackage{enumitem}
\usepackage{amsmath}
\usepackage{float}
\usepackage{multirow}

\title{Dimensionality and Measurement Precision in HLE's Multiple-Choice Subset}
\author{Mayank Sharma \\
Stanford University \\
\texttt{masharma@stanford.edu}
\And
Savira Nadela \\
Stanford University \\
\texttt{savira@stanford.edu}
\And
Tyler Matteson \\
Stanford University \\
\texttt{tylerjm@stanford.edu}
}

\begin{document}

\ifcolmsubmission
\linenumbers
\fi

\maketitle

\begin{abstract}
Humanity's Last Exam (HLE) is widely used to evaluate frontier language models. HLE organizes its questions into eight subject-domain categories, whose subscores are often interpreted as evidence of distinct capabilities. However, no study has assessed whether these labels correspond to empirically separable latent constructs, nor whether the benchmark effectively differentiates between models of similar ability. We evaluate 29 LLMs on the text-only multiple-choice subset of HLE and apply psychometric methods to assess both the dimensionality of the benchmark and the distribution of its measurement precision. 
Fitting a two-parameter logistic IRT model, we find that HLE is dominated by a single general reasoning factor: domain labels explain only 3.5\% of variance in the leading principal components of item responses, and within- and between-domain residual correlations are nearly identical (Cohen's $d = 0.016$). A simulation study with the same number of models ($N = 29$) shows that our analyses would detect clearly distinct domain abilities, so their absence is informative; however, small domain-specific differences cannot be ruled out, and model rankings vary across domains somewhat more than a single factor predicts. A separate analysis of the test information function reveals that measurement precision concentrates at moderate ability levels and drops sharply above $\theta = 0$, where the strongest models sit. These findings suggest that the subset's domain subscores do not warrant distinct capability interpretations and that its ability to discriminate among the strongest models might be limited.

\end{abstract}

\section{Introduction}

Humanity's Last Exam (HLE) has quickly emerged as a prominent benchmark for
evaluating advanced language models, appearing in capability assessments and
policy discussions shortly after its release \citep{Phan2026}. Designed to
resist simple retrieval and pattern matching through expert-level questions
spanning mathematics, the natural sciences, and the humanities, HLE represents
a plausible candidate for evaluating higher-order reasoning. However, widespread
adoption has outpaced systematic evaluation of its measurement properties. Most
benchmark studies, including HLE, report aggregate accuracy and domain-specific
subscores without testing whether the reported domains correspond to empirically
distinct latent factors. These subscores are often interpreted as evidence that
one model outperforms another in domains such as mathematics or chemistry, an
interpretation that implicitly assumes benchmark categories recover as separable
latent capabilities rather than as facets of a single general reasoning factor
\citep{Jiang2026}. Whether HLE's domain labels reflect psychometrically distinct
abilities remains largely untested, and the implications are concrete: if the
eight domains do not correspond to distinct latent constructs, then per-domain
model rankings reported in leaderboards are not warranted by the instrument, and
developers and policy audiences should treat HLE as a measure of general reasoning
rather than a profile of domain-specific skill.

A second concern is measurement precision. Even if HLE is unidimensional, its
ability to differentiate between models depends on where along the ability
continuum its items concentrate information. If most items are calibrated for
average-performing models, score differences between the strongest frontier
models may reflect measurement noise rather than genuine capability gaps, a
problem that will intensify as models continue to improve.

This paper addresses both concerns directly. We evaluate 29 LLMs on
the text-only MCQ subset of HLE ($J = 428$ items) and apply
psychometric methods to investigate two questions: (a) \textbf{Does HLE's
eight-domain structure reflect distinct latent constructs, or do the domains
collapse into a general reasoning factor?}, addressed through PCA of item response profiles, residual correlation analysis, inter-domain accuracy correlations, domain-level ability comparisons, and a simulation study of the statistical power of these analyses at $N = 29$; and (b) \textbf{Where
along the ability continuum does HLE concentrate measurement precision, and which
domains contribute most to discrimination among frontier models?}, addressed
through the test information function decomposed by subject domain.

\subsection{Related Work}

Benchmarks are the primary instrument by which progress in large language models
is evaluated, but their useful lifespan is often short. Popular benchmarks such
as MMLU have shown signs of saturation within only a few years of release
\citep{Phan2026}. This has motivated the development of HLE, a benchmark
consisting of 2,500 expert-level questions spanning mathematics, humanities, and
the natural sciences, crowdsourced from nearly 1,000 subject-matter experts
across 50 countries and filtered to ensure that contemporary language models
could not reliably answer them at submission time \citep{Phan2026}. HLE and
related studies (e.g., MMLU \citep{hendrycks2021mmlu}, GPQA
\citep{rein2023gpqagraduatelevelgoogleproofqa}) typically report aggregate
accuracy as the primary evaluation metric, but aggregate scores alone obscure
whether items meaningfully discriminate between models and whether domain
subscores reflect distinct latent constructs.

These measurement properties matter because benchmark rankings increasingly
inform high-stakes decisions ranging from model deployment to AI governance
\citep{liang2023helm}. When domain subscores are interpreted as evidence of
distinct capabilities, the validity of those comparisons is often assumed rather
than empirically demonstrated, and decisions based on poorly validated
measurements risk misrepresenting both relative model capability and the nature
of progress in language model development \citep{bommasani2022opportunitiesrisksfoundationmodels}.

Psychometric methods offer a framework for evaluating these properties directly.
Applying IRT across 29 NLP datasets, \citet{vania2021comparing}
showed that several widely used benchmarks contain items that fail to discriminate
effectively between models, raising questions about whether aggregate scores
reflect meaningful capability differences. TinyBenchmarks \citep{Polo2024} uses
IRT to enable efficient evaluation on subsets of MMLU, and Chatbot Arena
\citep{Chiang2024} applies Bradley-Terry models to pairwise comparisons. These
studies demonstrate the feasibility of treating benchmarks as measurement
instruments, but they focus on ranking efficiency and item difficulty rather than
on latent dimensionality: the question of whether domain-specific capability
claims are structurally warranted. More broadly, benchmark leaderboards
frequently report domain-specific performance as evidence of separable
capabilities without testing whether such categories recover as empirically
distinct latent factors \citep{Jiang2026}. To our knowledge, no psychometric
dimensionality analysis has yet been applied to HLE. This study addresses that
gap.

\section{Methods}

\subsection{HLE Subset}

HLE items contain two question formats: exact-match short-answer
($\approx$76\%) and multiple-choice ($\approx$24\%). Approximately 14\% of all
questions require image comprehension. The coverage of the subjects in the benchmark is as follows: mathematics (41\%), biology/medicine (11\%), computer science/AI (10\%),
physics (9\%), humanities/social science (9\%), other (9\%), chemistry (7\%) 
and engineering (4\%). We restrict our analysis to the \emph{text-only multiple-choice} subset,
which is the only subset amenable to automated binary scoring without multimodal
infrastructure or LLM-judge extraction. This comprises $J = 513$ items ($\approx$21\% of HLE),
with individual items containing between 2 and 21 answer choices. The full dataset
is publicly available via HuggingFace \footnote{https://huggingface.co/datasets/cais/hle} (\texttt{cais/hle}, \texttt{split="test"})
and is filtered to this subset by selecting \texttt{answer\_type == "multipleChoice"}
and excluding image-dependent items. See example items in Appendix A.1.

\subsection{Model Selection}

For generating responses on the subset, we used 37 contemporary language
models across five model families:
\textbf{OpenAI} ($n=12$): GPT-4.1 series (standard, mini, nano), GPT-5 series
(mini, nano, 5.4-mini, 5.4-nano), GPT-4o series (2024-11-20, mini), and reasoning
models (o3-mini, o4-mini, o1-mini);
\textbf{Anthropic} ($n=6$): Claude Opus (4.7, 4.6, 4.5), Claude Sonnet (4.6, 4.5),
and Claude Haiku 4.5;
\textbf{Google} ($n=5$): Gemini 3.5 Flash, Gemini 3.1 Flash Lite, Gemini 2.5 series
(Pro, Flash, Flash Lite);
\textbf{DeepSeek} ($n=4$): DeepSeek-V3, DeepSeek-V4 (Flash, Pro), and DeepSeek-R1
(reasoning);
\textbf{Open-weight models} ($n=10$): Gemma (2-9B, 3-27B), Phi-4, Qwen2.5-7B,
QwQ-32B (reasoning), Sky-T1-32B (reasoning), Mistral-7B, OLMo-2-7B, Falcon3-10B,
and Llama-3.1-8B. Models represented both reasoning-specialized architectures and
standard instruction-tuned models.
\definecolor{darkpurple}{RGB}{80, 50, 140}
\subsection{Response Collection}

For each model-item pair, we prompted models using a standardized format (see prompt in Appendix A.2). To ensure reproducibility, we set \texttt{temperature=0.0} for all models
supporting this parameter. Reasoning-specialized models (OpenAI o-series,
Claude Opus 4.7) used their default extended reasoning configurations without
temperature control. Maximum response lengths were set to 8,192 tokens, with model-specific reductions for models with shorter context windows (e.g., 4,096 tokens for Gemma-2-9B). Proprietary models (OpenAI, Anthropic, Google, DeepSeek) were queried via their
respective API endpoints with asynchronous request handling and automatic retry
logic for rate-limit management. Open-weight models were deployed using vLLM
\citep{kwon2023efficientmemorymanagementlarge} on cloud GPU infrastructure (Modal Labs) using
Nvidia A100-40GB, Nvidia A100-80GB, and Nvidia H100 GPUs. Data collection spanned
approximately 20 hours, with the majority of latency attributable to
reasoning-specialized models.

\subsection{Response Parsing and Scoring}

Model responses were parsed using rule-based extraction with a hierarchical
strategy: (1) regex pattern matching for explicit answer declarations
(``Answer:'', ``Final Answer''), supporting markdown formatting and parenthetical
notation (\texttt{(A)}); (2) flexible pattern matching for natural language
phrasings (``the correct answer is A,'' ``choice is B''); and (3) fallback
extraction of the last isolated letter (A-Z) when no explicit answer marker was
present. Parsed responses (91.8\%) were scored as correct (1) or incorrect (0)
by exact-match comparison to the ground-truth answer. Unparsed responses (8.2\%)
were coded as missing data and predominantly resulted from empty model outputs
(DeepSeek variants), failed 
API responses (Gemini 2.5 Pro), or inference
failures (Phi-4, Gemma-9B).

\subsection{Response Matrix Construction}

We constructed a binary response matrix $\mathbf{X} \in \{0, 1, \text{NA}\}^{N \times J}$ where $N$ represents models and $J=513$ items. From the initial 37 models evaluated, one model (o1-mini) was excluded due to zero coverage (0\% valid responses). Coverage among the remaining 36 models ranged from 50\% to 100\%, with median 99.7\% and mean 91.8\%. Because missingness was not at random (attributable to content safety filtering, inference failures, and rate limiting rather than item difficulty), we retained only the 29 models with excellent coverage ($\geq$95\%). Among the 513 items evaluated with these 29 models, 428 items (83.4\%) exhibited non-zero variance across models; the remaining 85 items (16.6\%) showed zero variance, with all models responding incorrectly, indicating extreme difficulty. We removed zero-variance items to obtain a final analytic matrix of $\mathbf{X} \in \{0, 1\}^{29 \times 428}$, comprising 12,412 model-item observations, with remaining sparse missing values (0.73\% of observations; $n=90$) coded as incorrect responses, as their negligible proportion introduces minimal bias to estimates.

A sensitivity analysis retaining all 36 models with nonzero coverage is described in Section~\ref{sec:sens-methods}.

\subsection{Analytic Overview}

First, we ask whether HLE's eight subject-domain labels correspond to empirically distinct latent constructs or collapse to a single general factor. Second, we ask how well HLE measures along the underlying factor(s): where measurement precision concentrates along the ability continuum, and which domains contribute most. We address both questions through a unified analysis, first estimating a two-parameter logistic (2PL) IRT model to obtain item-level discrimination and difficulty parameters, which serve as shared inputs to both stages, the dimensionality analysis and the measurement precision analysis.

\subsubsection{2PL model estimation}
We modeled item-level measurement properties using a two-parameter logistic (2PL)
IRT model \citep{baker2004item} fit to $\mathbf{X}$. The probability that model $i$ correctly answers
item $j$ was modeled as:
\begin{equation}
P(X_{ij}=1 \mid \theta_i,a_j,b_j)
=
\frac{1}{1+\exp[-a_j(\theta_i-b_j)]},
\end{equation}
where $\theta_i$ represents latent model ability, $a_j$ is the item discrimination
parameter, and $b_j$ is the item difficulty parameter. Higher discrimination values
indicate items that better differentiate between stronger and weaker models, whereas
higher difficulty values indicate items requiring greater latent ability for a 50\% probability of correct response. The model was estimated by joint maximum likelihood implemented in \texttt{torch\_measure} \citep{torch_measure}, with optimization run for up to 2{,}000 epochs using a learning rate of 0.05. Because joint estimation with $N = 29$ respondents produces unbounded discrimination estimates for items that nearly perfectly separate stronger from weaker models (14-26\% of items exceeded $a = 5$, depending on initialization), discrimination estimates were truncated at $a = 5$ after estimation for all downstream computations. Joint estimation with $N = 29$ is also sensitive to initialization, so, absolute parameter values (e.g., median $b$, the location of the TIF peak) shifted across random starts, but the qualitative patterns reported in the next sections (which domains were hardest and easiest, most and least discriminating, and most and least informative at the frontier, and the identity of the top-ranked models) were stable. For each item, we extracted estimated discrimination ($\hat{a}_j$) and difficulty ($\hat{b}_j$).

\subsubsection{Dimensionality analysis}

We examined whether the benchmark's 8 subject-domain labels reflected empirically distinct latent constructs. Using data from \(N = 29\) models and \(J = 428\) items, we found that the inter-item tetrachoric correlation matrix was rank-deficient and non-positive-definite, preventing the use of confirmatory factor analysis because the resulting fit indices would be invalid \citep{flora2004}.
We therefore used four $N$-robust alternatives that do not require inverting the correlation matrix, and evaluated the statistical power of each through simulation.

\paragraph{PCA on item response profiles.} We transposed the response matrix to $\mathbf{X}^\top \in \{0,1\}^{J \times N}$,
treating each item as a point in $N$-dimensional model space, and applied
standard PCA. If domain labels capture real structure, items from the same
domain should cluster together in this space. We quantified this using domain
$R^2$, the proportion of variance in the first three principal components
explained by domain membership. Because $R^2 > 0$ even under a single factor at finite $N$, we interpret it relative to its simulated null distribution.

\paragraph{Residual item correlations.} To further assess whether domain membership explains structure beyond the general factor, we converted the (capped) 2PL discriminations to approximate general-factor loadings, $\lambda_j = a_j / \sqrt{1 + a_j^2}$, computed Pearson correlations between item response vectors, subtracted the general-factor-implied correlation matrix $\mathbf{R}_g = \boldsymbol{\lambda}\boldsymbol{\lambda}^\top$, and compared within-domain to between-domain residual correlations using Cohen's $d$. If domains capture distinct latent structure, within-domain residuals should be higher than between-domain residuals. Because $\mathbf{R}_g$ and Pearson correlations between binary items are on different scales, we interpret $d$ relative to its simulated null distribution rather than its absolute size.

\paragraph{Domain-level $\hat\theta$ correlations.} To assess whether domain subscores carry information beyond the total score, we fit separate 2PL models within each of the four domains with at least 50 items and computed Pearson and Spearman correlations between domain-specific ability estimates $\hat{\theta}_{\text{domain}}$ and overall ability estimates $\hat{\theta}$. Because estimation noise keeps these correlations below 1 even under a single factor, we compare the smallest of the four Pearson correlations to its simulated null distribution.

\paragraph{Inter-domain accuracy correlations.} For each pair of domains, we computed the Spearman correlation, across models, between models' accuracies in the two domains, and took the median over all 28 pairs. Under a single factor, domains should rank models consistently, up to sampling noise within each domain.

\paragraph{Sensitivity analysis.} \label{sec:sens-methods} To assess whether excluding lower-coverage models affected the dimensionality results, we repeated the analyses with all 36 models with nonzero coverage (minimum coverage 50\%). Because missingness in several of these models was concentrated in particular domains, coding missing responses as incorrect could create spurious domain structure. We therefore treated missing responses as missing in all 2PL fits, which use only observed responses. PCA and Pearson correlations require a complete matrix, so for these, missing responses were replaced by the item’s mean accuracy; this procedure tends to attenuate correlations. After removing zero-variance items, the expanded matrix contained 36 models and 444 items.

\paragraph{Power simulation.} \label{sec:power-methods}
We simulated data to learn how each statistic behaves at our sample size ($N = 29$). Each simulated dataset contained 29 models and 428 items. Items reused the difficulty ($\hat b_j$) and discrimination ($\hat a_j$, capped at 5) estimated from the real data. Each simulated model was assigned $K$ abilities, one per factor, drawn from a normal distribution matching the mean and spread of the real $\hat\theta$ and correlated with each other at $\rho$. An item depended only on the ability for its domain's factor $f(j)$: the model answered correctly with probability $P(X_{ij}=1) = \sigma(a_j(\theta_{i,f(j)} - b_j))$, and each response was drawn as a random 0/1 outcome with that probability. We considered two-factor ($K=2$: Math, Physics, CS/AI, Engineering, and Chemistry vs.\ Biology/Medicine, Humanities/Social Science, and Other) and three-factor ($K=3$: Math, Physics, CS/AI, and Engineering vs.\ Biology/Medicine and Chemistry vs.\ Humanities/Social Science and Other) structures. In both cases, the inter-factor correlation $\rho$ took values $\{1, 0.9, 0.8, 0.7, 0.6, 0.5, 0.4, 0.3, 0.2, 0\}$, from 1 (a single ability) to 0 (fully distinct abilities).

Simulated datasets were subjected to the same analyses described above. Because the $\rho = 1$ condition is identical under both structures, it was simulated once. We simulated 4,100 datasets for domain $R^2$ and the inter-domain correlation (500 at $\rho = 1$ and 200 per $\rho < 1$ condition) and 2,100 for the more costly residual and domain-$\hat\theta$ analyses (300 and 100).

For each of the four statistics (domain $R^2$, residual $d$, inter-domain correlation, and minimum domain-$\hat\theta$ correlation), the $\rho = 1$ datasets show the values it takes when HLE measures a single ability, so that any apparent domain structure is due only to sampling noise. We call this its \emph{null distribution}. At $N = 29$ this noise is substantial: domain $R^2$, for example, averaged 0.031 across the null datasets rather than 0. Following the conventional 5\% significance level, we regarded a real-data value as evidence of domain structure only if fewer than 5\% of null datasets produced a value as extreme. ``Extreme'' means high for domain $R^2$ and residual $d$, and low for the two correlations. For domain $R^2$, this cutoff was 0.042; the real value of 0.035 fell below it. The $\rho < 1$ datasets, which contain domain abilities by design, show each statistic's \emph{power}: the proportion of datasets in which it passed its cutoff, that is, how often it detects domain structure that is really present. Power determines how a negative result should be read: if a high-power statistic does not pass its cutoff in the real data, domain structure is unlikely; if a low-power statistic does not, the result is inconclusive. Simulated single-ability data reproduced the real data's mean item--rest correlation (0.29 vs.\ 0.30) and split-half reliability (0.95 vs.\ 0.96), confirming that the null is a realistic baseline.

\subsubsection{Measurement precision}

Taking the 2PL estimates and dimensionality findings together, we characterize
where measurement precision concentrates along the ability continuum using the
test information function \citep{baker2004item},
\begin{equation}
I(\theta)
=
\sum_{j=1}^{J}
a_j^2
P_j(\theta)
\left[1-P_j(\theta)\right],
\end{equation}
which quantifies measurement precision at different ability levels, with higher
values corresponding to lower conditional SEM. We
decomposed the TIF by subject domain to evaluate which domains contributed most
strongly to discrimination.

\section{Results and Discussion}

\subsection{2PL Model Estimation}
\begin{figure}[htbp]
    \centering
    \includegraphics[width=0.8\textwidth]{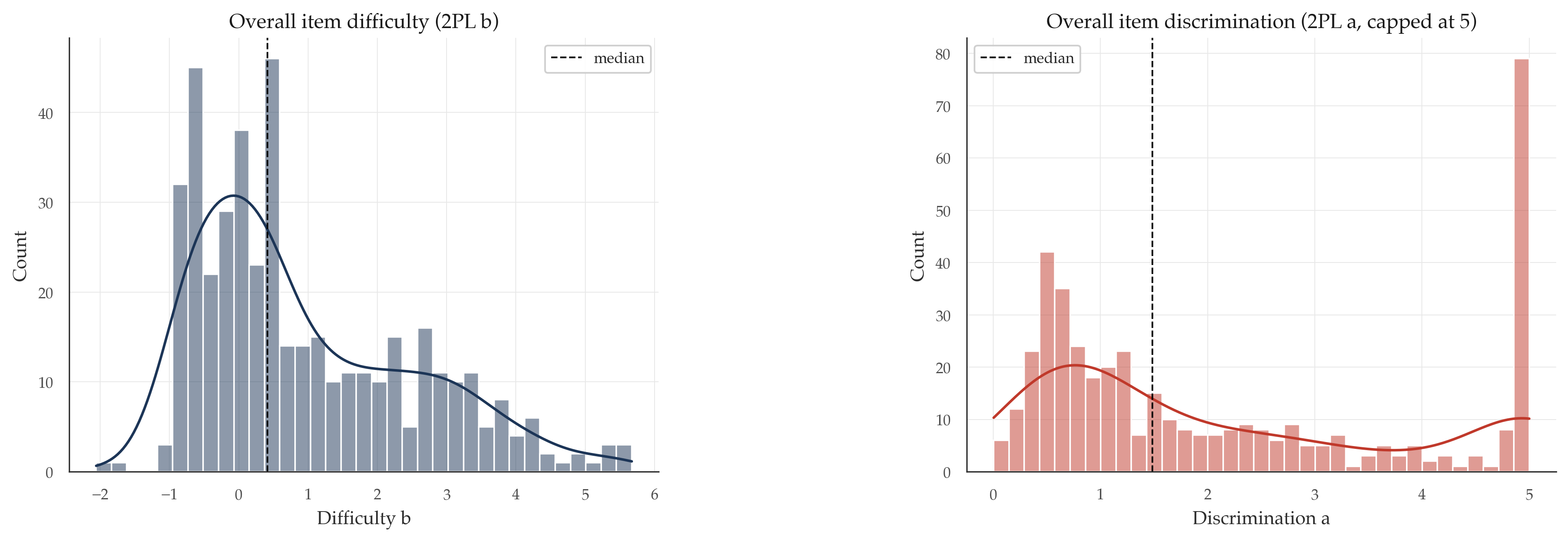}
    \caption{Distributions of item difficulty ($b$) and item discrimination
    ($a$).}
    \label{fig:overall_2pl}
\end{figure}

Across the full item pool, estimated difficulty parameters ranged from $-2.06$ to $5.67$, with a median of $b = 0.41$, indicating that the benchmark was moderately challenging for the evaluated models. The positively skewed distribution further suggests a substantial concentration of advanced items capable of differentiating performance across a wide range of contemporary LLMs \citep{liang2023helm}. Discrimination parameters ranged from $0.00$ to the imposed upper cap of $5.00$, with a median of $a = 1.49$. Many items demonstrated moderate-to-high discrimination
values, indicating that the benchmark effectively differentiated between stronger
and weaker models (see Figure~\ref{fig:overall_2pl}). Mean empirical accuracy across items was relatively low
($M = 0.17$), further supporting the conclusion that the benchmark was challenging for most models. See Appendix A.3 for supplemental results.

\begin{figure}[htbp]
    \centering
    \includegraphics[width=0.9\textwidth]{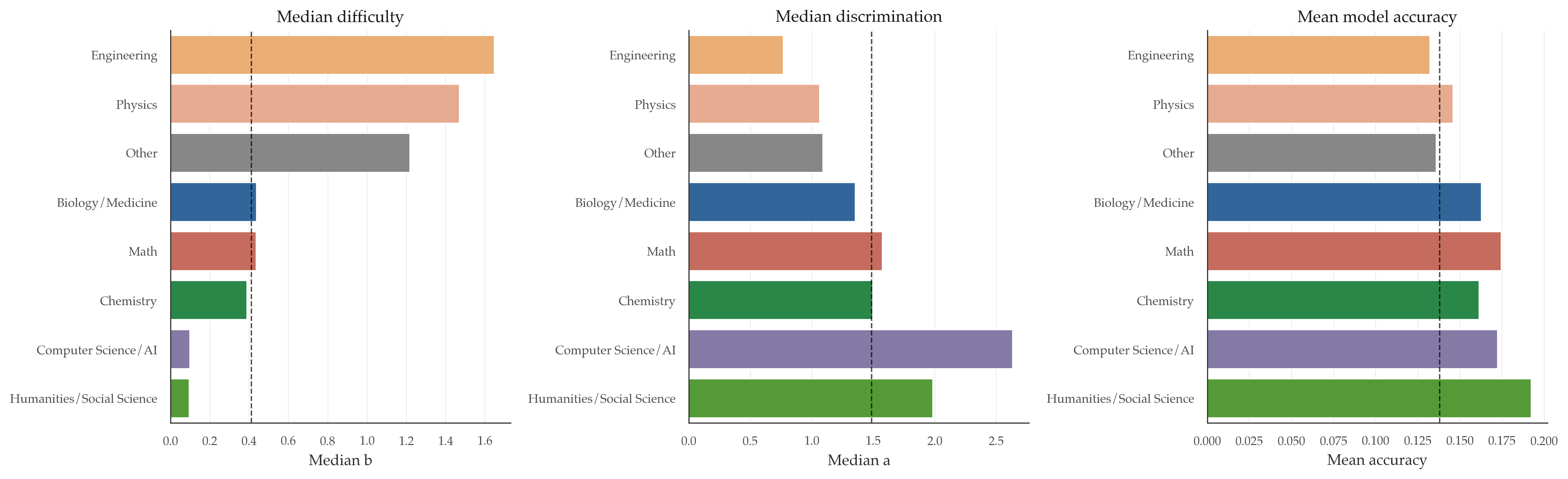}
    \caption{Category-level comparison of median item difficulty, median
    discrimination, and mean empirical accuracy across benchmark domains. Dashed
    vertical lines indicate overall median or mean values.}
    \label{fig:category_summary}
\end{figure}

\begin{figure}[H]
    \centering
    \includegraphics[width=\textwidth]{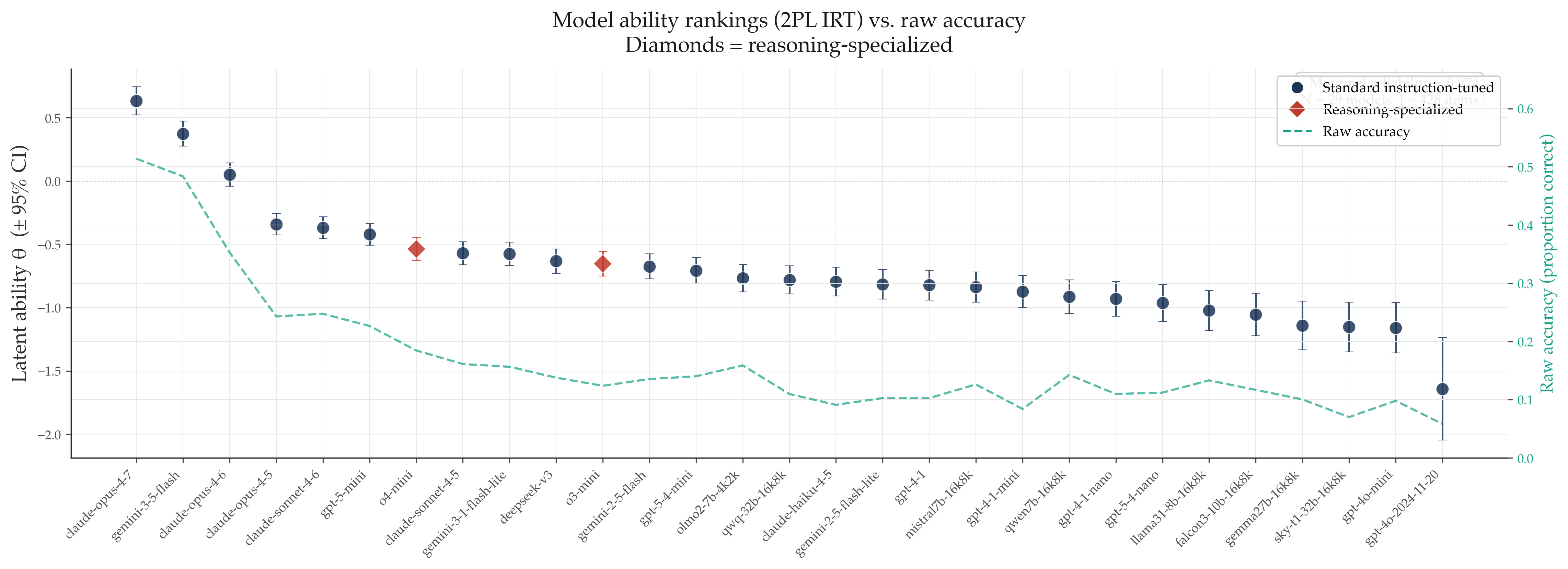}
   \caption{Estimated ability $\hat\theta$ for all 29 models with 95\% CIs (left axis), ordered by ability estimate, alongside raw
accuracy (right axis, dashed).
The two rankings are strongly correlated ($r_s = 0.857$).}
    \label{fig:ability_rankings}
\end{figure}

Substantial variation emerged across disciplinary categories (see Figure~\ref{fig:category_summary}). Engineering items exhibited the highest median difficulty ($b = 1.65$), whereas Computer Science/AI and Humanities/Social Science showed the lowest median difficulty values (both approximately $b = 0.10$). In contrast, discrimination patterns differed from difficulty trends. Computer Science/AI demonstrated the highest median discrimination ($a = 2.63$), while Engineering showed the lowest ($a = 0.77$), indicating that highly difficult domains did not necessarily provide the strongest differentiation between models. Category-level empirical accuracies aligned broadly with difficulty estimates, with Humanities/Social Science producing the highest mean accuracy ($M = 0.19$) and Engineering the lowest ($M = 0.13$). Figure~\ref{fig:ability_rankings} displays estimated latent ability $\hat\theta$
for all 29 models with 95\% confidence intervals alongside raw accuracy. Claude
Opus 4.7 achieved the highest estimated ability, followed by Gemini 3.5 Flash
and Claude Opus 4.6. GPT-4o-2024-11-20 exhibited an unusually low ability
estimate ($\hat\theta \approx -1.8$) with wide confidence intervals, suggesting
unstable parameter estimation likely due to atypically low or inconsistent
response patterns. IRT-based ability estimates were strongly correlated with
raw accuracy rankings ($r_s = 0.857$, $p < 10^{-8}$), confirming that the
2PL model recovers a latent dimension consistent with aggregate performance
while additionally characterizing item-level discrimination and measurement
precision unavailable from raw scores alone. However, category-level differences
in item parameters do not necessarily imply that HLE measures multiple distinct
latent abilities: domains could still reflect a single underlying reasoning
dimension if models that perform well in one domain tend to perform well across
all others. The following analysis examines this directly.

\subsection{Dimensionality Analysis}

\paragraph{PCA on item response profiles.}
Domain labels explained only $3.5\%$ of variance in the first three principal components (domain $R^2 = 0.035$). This value lies at the 75th percentile of its simulated unidimensional null (null mean $= 0.031$; 5\% critical value $= 0.042$), so it provides no evidence of domain clustering beyond what a single factor produces.

\paragraph{Residual item correlations.}
Within-domain and between-domain residual correlations were nearly identical
($\mu_\text{within} = -0.462$, $\mu_\text{between} = -0.466$, Cohen's $d = 0.016$).
This value lies at the 65th percentile of its simulated unidimensional null (null mean $= 0.006$; 5\% critical value $= 0.047$). Because the test has moderate power (0.55--0.72 at $\rho = 0.8$; 0.72--0.87 at $\rho = 0.7$), this result agrees with domain $R^2$ that items do not cluster by domain.\footnote{Residuals are negative because the general-factor-implied correlations $\lambda_j\lambda_k$ are on the latent-loading scale, whereas Pearson correlations between binary items are attenuated. Calibrating $d$ against simulated data processed identically accounts for this.}

\begin{figure}[t]
    \centering
    \includegraphics[width=\linewidth]{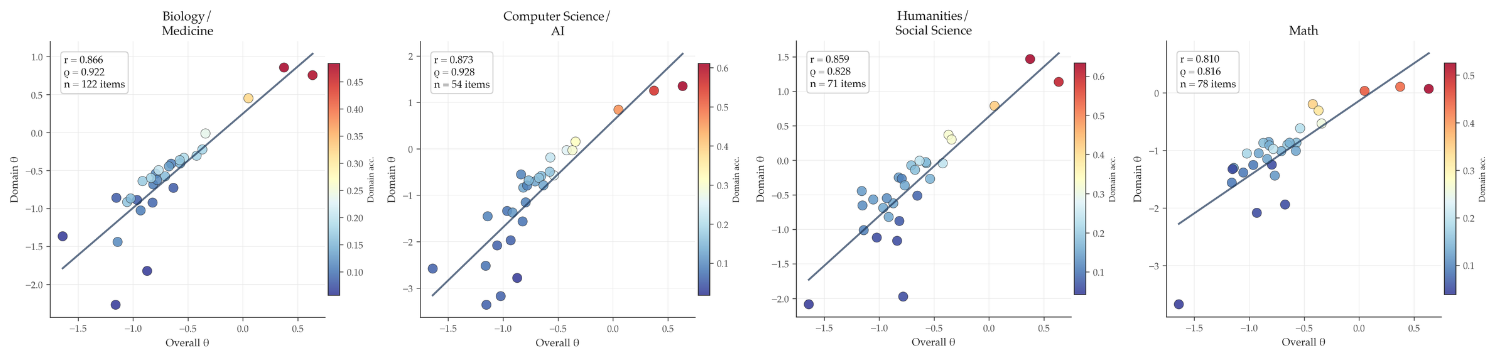}
    \caption{$\hat{\theta}_{\text{domain}}$ vs.\ $\hat{\theta}$ for four domains ($n \geq 50$). Each point is a model, colored by domain accuracy. $r \geq 0.81$, $r_s \geq 0.82$, consistent with a single factor at $N = 29$ (simulated null mean of the minimum $r$: 0.82).}
    \label{fig:subscale_theta}
\end{figure}

\paragraph{Domain-level $\hat\theta$ correlations.}
For the four domains with sufficient item counts ($n \geq 50$), domain $\hat\theta$
was strongly correlated with overall $\hat\theta$ (Figure~\ref{fig:subscale_theta}):
Computer Science/AI ($r = 0.873$, $r_s = 0.928$), Biology/Medicine
($r = 0.866$, $r_s = 0.922$), Humanities/Social Science ($r = 0.859$,
$r_s = 0.828$), and Mathematics ($r = 0.810$, $r_s = 0.816$; all $p < 10^{-7}$). Correlations for smaller domains (Engineering $n = 18$, Physics $n = 26$, Chemistry $n = 22$) were
substantially attenuated and are not interpreted, as fitting a 2PL to fewer than
30 items with $N = 29$ models yields poorly identified parameter estimates.
These correlations must be judged against what a single factor produces at this sample size, not against $r = 1$. In our simulation, the smallest of the four correlations averaged $0.82$ under perfect unidimensionality (90\% range $0.70$-$0.92$), so the observed values are as expected under one factor. However, this statistic has low power (about 0.20 at $\rho = 0.8$), and estimates varied across random initializations (minimum $r$ from $0.73$ to $0.84$ across 10 seeds). We therefore treat it as consistent with, but not strong evidence for, unidimensionality.

\paragraph{Inter-domain accuracy correlations.}
The median Spearman correlation between models' domain accuracies was $0.43$, lower than in all but 0.6\% of simulated unidimensional datasets (null mean $= 0.68$; $p \approx .006$). Model rankings therefore vary across domains more than a single factor predicts, even though items do not cluster by domain.

\subsection{Power Simulation}
\label{sec:power}

Table~\ref{tab:power} shows how often each statistic detected the domain structure built into the simulated data. At $\rho = 1$, where no domain structure exists, each statistic wrongly detected domain structure in about 5\% of datasets, as intended. Domain $R^2$ was the most sensitive, detecting structure in at least 79\% of datasets once $\rho \leq 0.8$, followed by residual $d$; the other two statistics needed more distinct abilities ($\rho \approx 0.5$--$0.6$). The simulations indicate that domain structures with \(\rho \leq 0.8\) would usually have been detected by the most sensitive analyses, whereas near-identical abilities (\(\rho=0.9\)) could not be reliably distinguished. The inter-domain correlation, despite its lower power, did reject the null ($0.43$, typical of $\rho \approx 0.5$--$0.6$). Thus, HLE's domains do not behave as separate abilities, although individual models' rankings shift somewhat from domain to domain.

\begin{table}[t]
\centering
\small
\setlength{\tabcolsep}{4pt}
\begin{tabular}{llcccc}
\toprule
 & & \multicolumn{4}{c}{Detection rate (two-factor / three-factor)} \\
\cmidrule(lr){3-6}
 & $\rho$ & domain $R^2$ & residual $d$ & inter-domain corr. & min $r(\hat\theta)$ \\
\midrule
False alarms & 1.0 & .05 / .05 & .05 / .05 & .05 / .05 & .05 / .05 \\
\midrule
\multirow{5}{*}{Power} & 0.9 & .37 / .33 & .26 / .37 & .13 / .14 & .09 / .12 \\
 & 0.8 & .79 / .79 & .55 / .72 & .26 / .31 & .21 / .17 \\
 & 0.7 & .97 / .97 & .72 / .87 & .50 / .56 & .37 / .57 \\
 & 0.5 & 1.0 / 1.0 & .84 / .97 & .80 / .97 & .74 / .85 \\
 & 0.0 & 1.0 / 1.0 & .75 / .97 & 1.0 / 1.0 & 1.0 / 1.0 \\
\midrule
\multicolumn{2}{l}{Real data: value} & .035 & .016 & .43 & .81 \\
\multicolumn{2}{l}{Real data: null percentile} & 75th & 65th & 0.6th & 36th \\
\bottomrule
\end{tabular}
\caption{Detection rates for each statistic across 29-model simulated datasets ($J = 428$), at $\alpha = 0.05$, for two-factor / three-factor structures. At $\rho = 1$ no domain structure exists, so detections are false alarms. Selected values of $\rho$ are shown; all values are plotted in Appendix~\ref{app:power}. Bottom rows: the real-data value of each statistic and its percentile among the $\rho = 1$ datasets. Only the inter-domain correlation falls in the extreme 5\%.}
\label{tab:power}
\end{table}

\subsection{Sensitivity analysis} Results were similar under the relaxed coverage threshold. In the 36-model analysis, domain $R^2 = 0.043$ and the within- versus between-domain residual difference remained small ($d = 0.028$). Domain-specific ability estimates remained strongly rank-correlated with overall ability across well-powered domains ($r_s = 0.834$-$0.919$), although the Pearson correlation for Humanities/Social Science was lower ($r = 0.645$). These values were not recalibrated against a simulated null at $N = 36$, so we treat them as a qualitative robustness check: including lower-coverage models did not reveal domain structure absent from the main analysis.

\subsection{Measurement Precision}

\begin{figure}[t]
    \centering
    \includegraphics[width=0.7\linewidth]{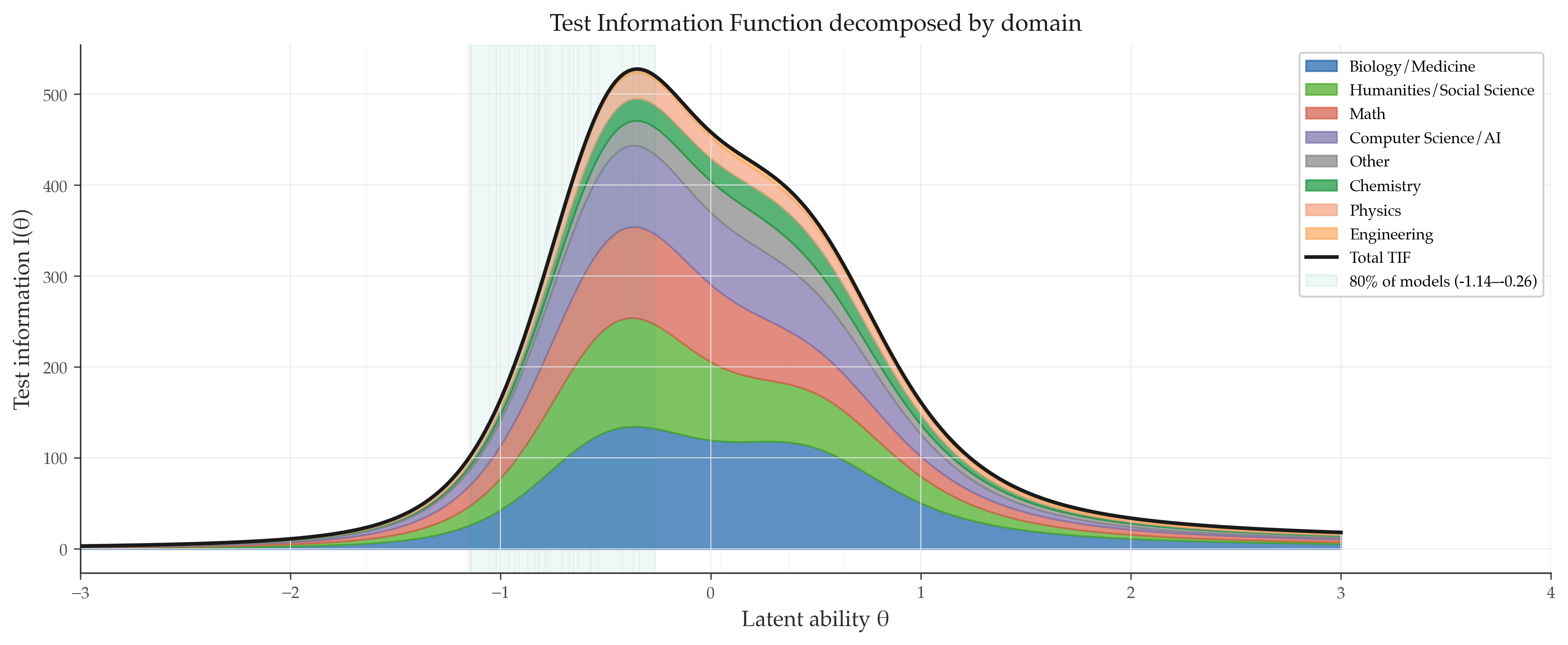}
    \caption{TIF decomposed by domain (stacked
    area). The shaded band marks the 10th-90th percentile ability range of
    models ($\theta \in [-1.14, -0.26]$), showing that HLE concentrates
    measurement precision at moderate ability levels. The TIF drops sharply
    above $\theta = 0$}
    \label{fig:tif_domain}
\end{figure}

The test information function (TIF) peaks at $\theta = -0.35$, coinciding with
the ability range of most models in our sample (10th--90th percentile interval:
$[-1.14, -0.26]$). Measurement precision drops substantially above $\theta = 0$, where only the two or three strongest models in our sample sit, and where future frontier models will be located. Decomposing the TIF by domain (Figure~\ref{fig:tif_domain}), Biology/Medicine contributes the most total
information in absolute terms ($I = 12{,}611$). However, Engineering concentrates
the highest proportion of its information at the frontier ($69.1\%$ at $\theta > 0$),
consistent with its extreme item difficulty. Mathematics and Humanities/Social
Science show the lowest frontier concentration ($41.5\%$ and $41.4\%$ respectively),
and despite containing more items than most other domains ($n = 78$ and $n = 71$), contribute disproportionately little to discrimination among higher-performing
models.

\section{Conclusion and Future Work}
Fitting a 2PL IRT model to responses from 29 language models on HLE's
text-only multiple-choice subset ($J = 428$ items), we find that the benchmark is dominated by a single general reasoning factor. Domain labels explained only 3.5\% of variance in the leading principal components of item responses, residual correlations after removing the general factor were nearly identical within and between domains (Cohen's $d = 0.016$), and domain-specific ability estimates tracked the total score as closely as expected under a single factor. A power simulation calibrated to our sample shows that these data make separate domain factors (correlated at $\rho \leq 0.8$) unlikely but cannot exclude weaker ones, and inter-domain accuracy correlations suggest some domain-specific variance in model rankings. HLE's domain labels therefore do not support claims of distinct capabilities, although small domain-specific differences cannot be excluded at this sample size. A
separate finding concerns measurement precision. While HLE discriminates well
among models at moderate ability levels, precision drops substantially above
$\theta = 0$, where the strongest models in our sample sit. Engineering items concentrate
the most information at the frontier ($69.1\%$ at $\theta > 0$) but represent
only 4\% of the benchmark; Mathematics and Humanities/Social Science, the two
largest domains, contribute disproportionately little to frontier discrimination.
As stronger models continue to emerge, this precision gap will become an
increasingly binding constraint on HLE's utility as a differentiating instrument.

The psychometric approach was informative, but only once each statistic was calibrated against a simulated null at the actual sample size. This calibration showed that domain $R^2$ and residual correlations are reasonably powered, and that domain-$\hat\theta$ correlations near 0.8 are the expected value under a single factor rather than evidence of redundancy. The primary methodological
challenge was the small model sample ($N = 29$), which precluded confirmatory
factor analysis and limited statistical power for domain-level analyses. This
inverted the usual psychometric setting, where items are typically far fewer
than subjects, and required N-robust alternatives throughout. A practical lesson
for future benchmark psychometrics is that model coverage must be treated as a
first-class design consideration: missingness that is not at random, as observed here, can introduce selection bias and limits generalizability.

Several limitations bound these conclusions. Our analytic sample is restricted to the text-only multiple-choice subset ($\approx$21\% of HLE), so findings do not extend to exact-match or image-dependent items; moreover, the multiple-choice format may introduce construct-irrelevant variance if models succeed through elimination rather than genuine domain knowledge. The model sample is small ($N = 29$) and non-random, although a sensitivity analysis retaining all 36 models with nonzero coverage produced substantively similar dimensionality results. Strongly domain-dependent missingness nevertheless limits generalizability, and the small, non-random model sample precludes formal measurement invariance testing across model families. Scoring each model on a single deterministic response per item means discrimination estimates reflect stable model differences but not within-model response variability, which may underestimate uncertainty in item parameter estimates. Our power simulation considered only simple structures aligned with HLE's domain labels; factors that cut across domains would not be detected by domain-based statistics, and generating parameters were estimated from the same small sample. Future work should replicate on the full benchmark with a larger and more complete model sample to enable CFA and invariance testing.

\section*{Code Availability}
All analysis code used in this study is available at: \url{https://github.com/matrix-mayank/hle-psychometrics}.

\section*{AI Usage}

AI tools were used in a limited way throughout this project, mainly to help improve writing clarity, troubleshoot coding issues, and brainstorm ways to present results more clearly. To avoid plagiarism or non-attribution of ideas, all citations and references were added manually by the authors after checking the original academic sources directly. We also reviewed AI-generated suggestions carefully to make sure ideas, interpretations, and wording were appropriately credited and not copied without attribution. Responsibility for the originality, accuracy, and scientific integrity of the final paper remained entirely with the authors.

\section*{Ethics Statement}
This project examines the psychometric properties of a large language model benchmark and highlights the importance of carefully interpreting benchmark scores and domain subscores. One potential positive impact is encouraging more rigorous evaluation of benchmark validity, especially when benchmark rankings are used in research, industry, or policy discussions. However, the findings could also be misinterpreted if generalized beyond the analyzed subset or treated as definitive statements about model intelligence. To reduce this risk, the paper clearly discusses limitations such as the small model sample and restriction to the multiple-choice subset of HLE.

\bibliography{references}

@article{Phan2026,
  author = {Phan, Long and Gatti, Alice and Li, Nathaniel and Khoja, Adam and Kim, Ryan and Ren, Richard and Hausenloy, Jason and Zhang, Oliver and Mazeika, Mantas and Hendrycks, Dan and Han, Ziwen and Hu, Josephina and Zhang, Hugh and Zhang, Chen Bo Calvin and Shaaban, Mohamed and Ling, John and Shi, Sean and Choi, Michael and Agrawal, Anish and Chopra, Arnav and Nattanmai, Aakaash and McKellips, Gordon and Cheraku, Anish and Suhail, Asim and Luo, Ethan and Deng, Marvin and Luo, Jason and Zhang, Ashley and Jindel, Kavin and Paek, Jay and Halevy, Kasper and Baranov, Allen and Liu, Michael and Avadhanam, Advaith and Zhang, David and Cheng, Vincent and Ma, Brad and Fu, Evan and Do, Liam and Lass, Joshua and Yang, Hubert and Sunkari, Surya and Bharath, Vishruth and Ai, Violet and Leung, James and Agrawal, Rishit and Zhou, Alan and Chen, Kevin and Kalpathi, Tejas and Xu, Ziqi and Wang, Gavin and Xiao, Tyler and Maung, Erik and Lee, Sam and Yang, Ryan and Yue, Roy and Zhao, Ben and Yoon, Julia and Sun, Xiangwan and Singh, Aryan and Peng, Clark and Osbey, Tyler and Wang, Taozhi and Echeazu, Daryl and Wu, Timothy and Patel, Spandan and Kulkarni, Vidhi and Sundarapandiyan, Vijaykaarti and Le, Andrew and Nasim, Zafir and Yalam, Srikar and Kasamsetty, Ritesh and Samal, Soham and Sun, David and Shah, Nihar and Saha, Abhijeet and Zhang, Alex and Nguyen, Leon and Nagumalli, Laasya and Wang, Kaixin and Wu, Aidan and Telluri, Anwith and Yue, Summer and Wang, Alexandr and Dodonov, Dmitry and Nguyen, Tung and Lee, Jaeho and Anderson, Daron and Doroshenko, Mikhail and Stokes, Alun Cennyth and Mahmood, Mobeen and Pokutnyi, Oleksandr and Iskra, Oleg and Wang, Jessica P. and Levin, John-Clark and Kazakov, Mstyslav and Feng, Fiona and Feng, Steven Y. and Zhao, Haoran and Yu, Michael and Gangal, Varun and Zou, Chelsea and Wang, Zihan and Popov, Serguei and Gerbicz, Robert and Galgon, Geoff and Schmitt, Johannes and Yeadon, Will and Lee, Yongki and Sauers, Scott and Sanchez, Alvaro and Giska, Fabian and Roth, Marc and Riis, S{\o}ren and Utpala, Saiteja and Burns, Noah and Goshu, Gashaw M. and Naiya, Mohinder Maheshbhai and Agu, Chidozie and Giboney, Zachary and Cheatom, Antrell and Fournier-Facio, Francesco and Crowson, Sarah-Jane and Finke, Lennart and Cheng, Zerui and Zampese, Jennifer and Hoerr, Ryan G. and Nandor, Mark and Park, Hyunwoo and Gehrunger, Tim and Cai, Jiaqi and McCarty, Ben and Garretson, Alexis C. and Taylor, Edwin and Sileo, Damien and Ren, Qiuyu and Qazi, Usman and Li, Lianghui and Nam, Jungbae and Wydallis, John B. and Arkhipov, Pavel and Shi, Jack Wei Lun and Bacho, Aras and Willcocks, Chris G. and Cao, Hangrui and Motwani, Sumeet and de Oliveira Santos, Emily and Veith, Johannes and Vendrow, Edward and Cojoc, Doru and Zenitani, Kengo and Robinson, Joshua and Tang, Longke and Li, Yuqi and Vendrow, Joshua and Fraga, Natanael Wildner and Kuchkin, Vladyslav and Maksimov, Andrey Pupasov and Marion, Pierre and Efremov, Denis and Lynch, Jayson and Liang, Kaiqu and Mikov, Aleksandar and Gritsevskiy, Andrew and Guillod, Julien and Demir, G{\"o}zdenur and Martinez, Dakotah and Pageler, Ben and Zhou, Kevin and Soori, Saeed and Press, Ori and Tang, Henry and Rissone, Paolo and Green, Sean R. and Br{\"u}ssel, Lina and Twayana, Moon and Dieuleveut, Aymeric and Imperial, Joseph Marvin and Prabhu, Ameya and Yang, Jinzhou and Crispino, Nick and Rao, Arun and Zvonkine, Dimitri and Loiseau, Gabriel and Kalinin, Mikhail and Lukas, Marco and Manolescu, Ciprian and Stambaugh, Nate and Mishra, Subrata and Hogg, Tad and Bosio, Carlo and Coppola, Brian P. and Salazar, Julian and Jin, Jaehyeok and Sayous, Rafael and Ivanov, Stefan and Schwaller, Philippe and Senthilkumar, Shaipranesh and Bran, Andres M. and Algaba, Andres and Van den Houte, Kelsey and Van Der Sypt, Lynn and Verbeken, Brecht and Noever, David and Kopylov, Alexei and Myklebust, Benjamin and Li, Bikun and Schut, Lisa and Zheltonozhskii, Evgenii and Yuan, Qiaochu and Lim, Derek and Stanley, Richard and Yang, Tong and Maar, John and Wykowski, Julian and Oller, Mart and Sahu, Anmol and Ardito, Cesare Giulio and Hu, Yuzheng and Kamdoum, Ariel Ghislain Kemogne and Jin, Alvin and Vilchis, Tobias Garcia and Zu, Yuexuan and Lackner, Martin and Koppel, James and Sun, Gongbo and Antonenko, Daniil S. and Chern, Steffi and Zhao, Bingchen and Arsene, Pierrot and Cavanagh, Joseph M. and Li, Daofeng and Shen, Jiawei and Crisostomi, Donato and Zhang, Wenjin and Dehghan, Ali and Ivanov, Sergey and Perrella, David and Kaparov, Nurdin and Zang, Allen and Sucholutsky, Ilia and Kharlamova, Arina and Orel, Daniil and Poritski, Vladislav and Ben-David, Shalev and Berger, Zachary and Whitfill, Parker and Foster, Michael and Munro, Daniel and Ho, Linh and Sivarajan, Shankar and Hava, Dan Bar and Kuchkin, Aleksey and Holmes, David and Rodriguez-Romero, Alexandra and Sommerhage, Frank and Zhang, Anji and Moat, Richard and Schneider, Keith and Kazibwe, Zakayo and Clarke, Don and Kim, Dae Hyun and Dias, Felipe Meneguitti and Fish, Sara and Elser, Veit and Kreiman, Tobias and Vilchis, Victor Efren Guadarrama and Klose, Immo and Anantheswaran, Ujjwala and Zweiger, Adam and Rawal, Kaivalya and Li, Jeffery and Nguyen, Jeremy and Daans, Nicolas and Heidinger, Haline and Radionov, Maksim and Rozho{\v{n}}, V{\'a}clav and Ginis, Vincent and Stump, Christian and Cohen, Niv and Po{\'s}wiata, Rafa{\l} and Tkadlec, Josef and Goldfarb, Alan and Wang, Chenguang and Padlewski, Piotr and Barzowski, Stanislaw and Montgomery, Kyle and Stendall, Ryan and Tucker-Foltz, Jamie and Stade, Jack and Rogers, T. Ryan and Goertzen, Tom and Grabb, Declan and Shukla, Abhishek and Givr{\'e}, Alan and Ambay, John Arnold and Sen, Archan and {Center for AI Safety} and {Scale AI} and {HLE Contributors Consortium}},
  title = {A benchmark of expert-level academic questions to assess AI capabilities},
  journal = {Nature},
  year = {2026},
  volume = {649},
  number = {8099},
  pages = {1139--1146},
  doi = {10.1038/s41586-025-09962-4},
  url = {https://doi.org/10.1038/s41586-025-09962-4},
  issn = {1476-4687}
}

@article{Polo2024,
  author = {Polo, Felipe Maia and Weber, Lucas and Choshen, Leshem and Sun, Yuekai and Xu, Gongjun and Yurochkin, Mikhail},
  title = {tinyBenchmarks: evaluating LLMs with fewer examples},
  journal = {arXiv preprint arXiv:2402.14992},
  year = {2024},
  url = {https://arxiv.org/abs/2402.14992}
}

@article{Chiang2024,
  author = {Chiang, Wei-Lin and Zheng, Lianmin and Sheng, Ying and Angelopoulos, Anastasios Nikolas and Li, Tianle and Li, Dacheng and Zhu, Banghua and Zhang, Hao and Jordan, Michael I. and Gonzalez, Joseph E. and Stoica, Ion},
  title = {Chatbot Arena: An Open Platform for Evaluating LLMs by Human Preference},
  journal = {arXiv preprint arXiv:2403.04132},
  year = {2024},
  url = {https://arxiv.org/abs/2403.04132}
}

@misc{rein2023gpqagraduatelevelgoogleproofqa,
      title={GPQA: A Graduate-Level Google-Proof Q\&A Benchmark}, 
      author={David Rein and Betty Li Hou and Asa Cooper Stickland and Jackson Petty and Richard Yuanzhe Pang and Julien Dirani and Julian Michael and Samuel R. Bowman},
      year={2023},
      eprint={2311.12022},
      archivePrefix={arXiv},
      primaryClass={cs.AI},
      url={https://arxiv.org/abs/2311.12022}, 
}

@article{hendrycks2021mmlu,
  title     = {Measuring Massive Multitask Language Understanding},
  author    = {Hendrycks, Dan and Burns, Collin and Basart, Steven and Zou, Andy and
               Mazeika, Mantas and Song, Dawn and Steinhardt, Jacob},
  journal   = {arXiv preprint arXiv:2009.03300},
  year      = {2021},
  eprint    = {2009.03300},
  archivePrefix = {arXiv},
  primaryClass  = {cs.CY},
  url       = {https://arxiv.org/abs/2009.03300}
}

@article{liang2023helm,
  title     = {Holistic Evaluation of Language Models},
  author    = {Liang, Percy and Bommasani, Rishi and Lee, Tony and Tsipras, Dimitris and
               Soylu, Dilara and Yasunaga, Michihiro and Zhang, Yian and Narayanan, Deepak
               and Wu, Yuhuai and Kumar, Ananya and others},
  journal   = {Transactions on Machine Learning Research},
  year      = {2023},
  eprint    = {2211.09110},
  archivePrefix = {arXiv},
  primaryClass  = {cs.CL},
  url       = {https://arxiv.org/abs/2211.09110}
}

@inproceedings{vania2021comparing,
    title = "Comparing Test Sets with Item Response Theory",
    author = "Vania, Clara and
      Htut, Phu Mon and
      Huang, William and
      Mungra, Dhara and
      Pang, Richard Yuanzhe and
      Phang, Jason and
      Liu, Haokun and
      Cho, Kyunghyun and
      Bowman, Samuel R.",
    booktitle = "Proceedings of the 59th Annual Meeting of the Association for Computational Linguistics and the 11th International Joint Conference on Natural Language Processing (Volume 1: Long Papers)",
    month = aug,
    year = "2021",
    address = "Online",
    publisher = "Association for Computational Linguistics",
    pages = "1141--1158",
    doi = "10.18653/v1/2021.acl-long.92",
    url = "https://aclanthology.org/2021.acl-long.92/"
}

@misc{bommasani2022opportunitiesrisksfoundationmodels,
      title={On the Opportunities and Risks of Foundation Models}, 
      author={Rishi Bommasani and Drew A. Hudson and Ehsan Adeli and Russ Altman and Simran Arora and Sydney von Arx and Michael S. Bernstein and Jeannette Bohg and Antoine Bosselut and Emma Brunskill and Erik Brynjolfsson and Shyamal Buch and Dallas Card and Rodrigo Castellon and Niladri Chatterji and Annie Chen and Kathleen Creel and Jared Quincy Davis and Dora Demszky and Chris Donahue and Moussa Doumbouya and Esin Durmus and Stefano Ermon and John Etchemendy and Kawin Ethayarajh and Li Fei-Fei and Chelsea Finn and Trevor Gale and Lauren Gillespie and Karan Goel and Noah Goodman and Shelby Grossman and Neel Guha and Tatsunori Hashimoto and Peter Henderson and John Hewitt and Daniel E. Ho and Jenny Hong and Kyle Hsu and Jing Huang and Thomas Icard and Saahil Jain and Dan Jurafsky and Pratyusha Kalluri and Siddharth Karamcheti and Geoff Keeling and Fereshte Khani and Omar Khattab and Pang Wei Koh and Mark Krass and Ranjay Krishna and Rohith Kuditipudi and Ananya Kumar and Faisal Ladhak and Mina Lee and Tony Lee and Jure Leskovec and Isabelle Levent and Xiang Lisa Li and Xuechen Li and Tengyu Ma and Ali Malik and Christopher D. Manning and Suvir Mirchandani and Eric Mitchell and Zanele Munyikwa and Suraj Nair and Avanika Narayan and Deepak Narayanan and Ben Newman and Allen Nie and Juan Carlos Niebles and Hamed Nilforoshan and Julian Nyarko and Giray Ogut and Laurel Orr and Isabel Papadimitriou and Joon Sung Park and Chris Piech and Eva Portelance and Christopher Potts and Aditi Raghunathan and Rob Reich and Hongyu Ren and Frieda Rong and Yusuf Roohani and Camilo Ruiz and Jack Ryan and Christopher Ré and Dorsa Sadigh and Shiori Sagawa and Keshav Santhanam and Andy Shih and Krishnan Srinivasan and Alex Tamkin and Rohan Taori and Armin W. Thomas and Florian Tramèr and Rose E. Wang and William Wang and Bohan Wu and Jiajun Wu and Yuhuai Wu and Sang Michael Xie and Michihiro Yasunaga and Jiaxuan You and Matei Zaharia and Michael Zhang and Tianyi Zhang and Xikun Zhang and Yuhui Zhang and Lucia Zheng and Kaitlyn Zhou and Percy Liang},
      year={2022},
      eprint={2108.07258},
      archivePrefix={arXiv},
      primaryClass={cs.LG},
      url={https://arxiv.org/abs/2108.07258}, 
}

@misc{kwon2023efficientmemorymanagementlarge,
      title={Efficient Memory Management for Large Language Model Serving with PagedAttention}, 
      author={Woosuk Kwon and Zhuohan Li and Siyuan Zhuang and Ying Sheng and Lianmin Zheng and Cody Hao Yu and Joseph E. Gonzalez and Hao Zhang and Ion Stoica},
      year={2023},
      eprint={2309.06180},
      archivePrefix={arXiv},
      primaryClass={cs.LG},
      url={https://arxiv.org/abs/2309.06180}, 
}

@article{Jiang2026,
  author = {Jiang, Han and Zhang, Susu and Zhu, Dongyao and Bai, Yuzhuo and Truong, Sang T. and Yi, Xiaoyuan and Koyejo, Sanmi and Xie, Xing and Xiao, Ziang},
  title = {AI Evaluation Should Require Standardized Item-Level Data Releases},
  journal = {arXiv preprint arXiv:2604.03244},
  year = {2026},
  url = {https://arxiv.org/abs/2604.03244}
}

@article{flora2004,
  author  = {Flora, David B. and Curran, Patrick J.},
  title   = {An empirical evaluation of alternative methods of estimation 
             for confirmatory factor analysis with ordinal data},
  journal = {Psychological Methods},
  year    = {2004},
  volume  = {9},
  number  = {4},
  pages   = {466--491},
  pmid    = {15598100},
  pmc     = {PMC3153362},
  doi     = {10.1037/1082-989X.9.4.466}
}

@software{torch_measure,
  author  = {Truong, Sang T. and others},
  title   = {torch\_measure: A Package for AI Measurement Science},
  year    = {2026},
  url     = {https://github.com/aims-foundations/torch_measure},
  note    = {MIT License}
}

@book{baker2004item,
  author    = {Baker, Frank B. and Kim, Seock-Ho},
  title     = {Item Response Theory: Parameter Estimation Techniques},
  edition   = {2},
  publisher = {CRC Press},
  year      = {2004},
  doi       = {10.1201/9781482276725}
}
\bibliographystyle{colm2026_conference}


\newpage

\appendix

\section{Appendix}

\subsection{Example Items from HLE}
\definecolor{darkpurple}{RGB}{85, 26, 139}
\begin{tcolorbox}[
    enhanced,
    breakable,
    colback=darkpurple!8,
    colframe=darkpurple!90,
    colbacktitle=darkpurple!90,
    coltitle=white,
    fonttitle=\bfseries\small,
    title={Example Items from HLE},
    label=box:hle-examples,
    arc=4pt,
    boxrule=1pt,
    left=8pt, right=8pt, top=6pt, bottom=6pt,
    titlerule=0pt,
    toptitle=4pt,
    bottomtitle=4pt,
]
\small

\begin{enumerate}[leftmargin=*, labelsep=6pt]
    \setlength\itemsep{8pt}

    \item \textbf{\textcolor{darkpurple!99}{Philosophy:}}\enspace ``Which condition
    of Arrhenius's sixth impossibility theorem do critical-level views violate?''
    with answer choices including Egalitarian Dominance, General Non-Extreme Priority,
    Non-Elitism, Weak Non-Sadism, and Weak Quality Addition.

    \item \textbf{\textcolor{darkpurple!99}{Linguistics:}}\enspace ``What is the
    standard Japanese pitch accent pattern of the word for `younger brother' in
    Japanese?'' with choices spanning Heiban, Atamadaka, Nakadaka, Odaka, and
    Heiban/Nakadaka.

\end{enumerate}
\end{tcolorbox}

\subsection{Response Collection Prompt}

\definecolor{darkturquoise}{RGB}{0, 115, 115}
\begin{tcolorbox}[
    enhanced,
    breakable,
    colback=darkpurple!5,
    colframe=darkpurple!70,
    colbacktitle=darkpurple!70,
    coltitle=white,
    fonttitle=\bfseries\small,
    title={Standardized Prompt Format},
    label=box:prompt-format,
    arc=4pt,
    boxrule=1pt,
    left=8pt, right=8pt, top=6pt, bottom=6pt,
    titlerule=0pt,
    toptitle=4pt,
    bottomtitle=4pt,
]
\small
\textit{Your response should be in the following format:}

\texttt{Explanation:} \textit{\{your explanation for your answer choice\}}\\
\texttt{Answer:} \textit{\{your chosen answer\}}\\
\texttt{Confidence:} \textit{\{your confidence score between 0\% and 100\% for
your answer\}}
\end{tcolorbox}

\subsection{Additional 2PL Diagnostic Plots}

\begin{figure}[h]
    \centering
    \includegraphics[width=\linewidth]{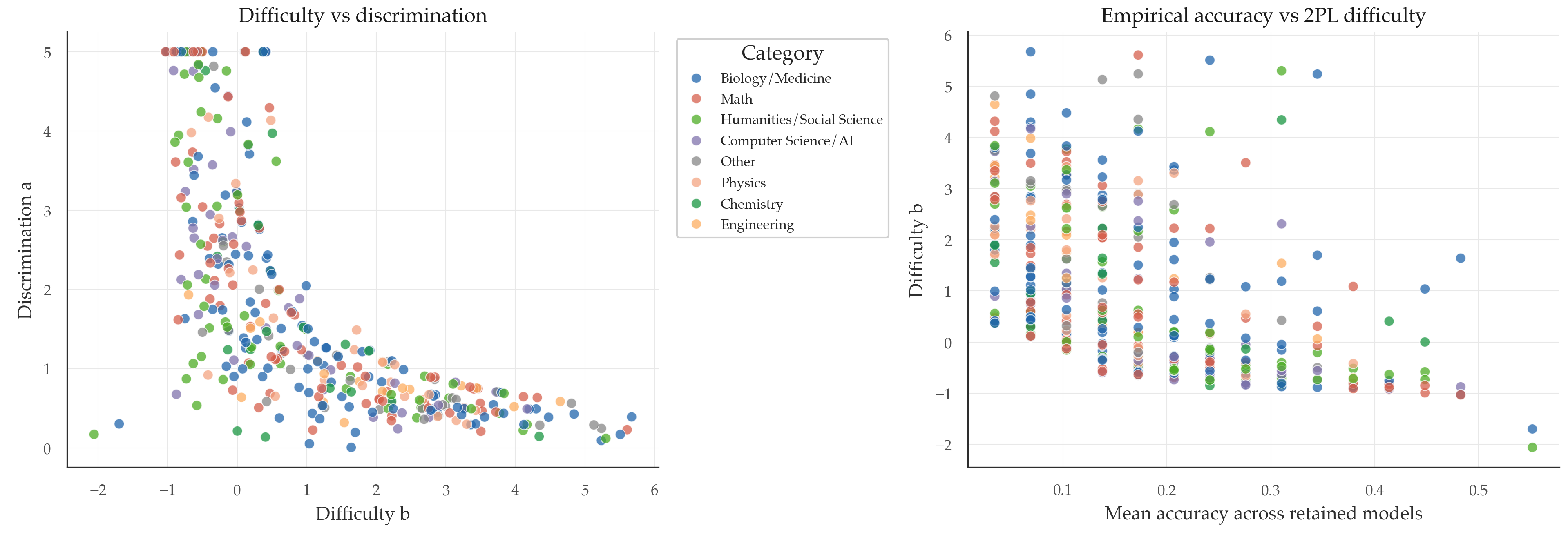}
    \caption{Supplementary 2PL diagnostic plots. Left: relationship between item difficulty and discrimination across benchmark domains. Right: empirical item accuracy against estimated 2PL difficulty.}
    \label{fig:appendix_2pl}
\end{figure}

Figure~\ref{fig:appendix_2pl} presents supplementary diagnostic visualizations for the estimated 2PL parameters. The left panel illustrates the inverse relationship between item difficulty and discrimination, where extremely difficult items tended to exhibit lower discrimination values. The right panel shows the expected negative relationship between empirical accuracy and estimated item difficulty, supporting the consistency of the estimated IRT parameters.

\newpage
\subsection{Power Curves}
\label{app:power}

\begin{figure}[h]
\centering
\includegraphics[width=\linewidth]{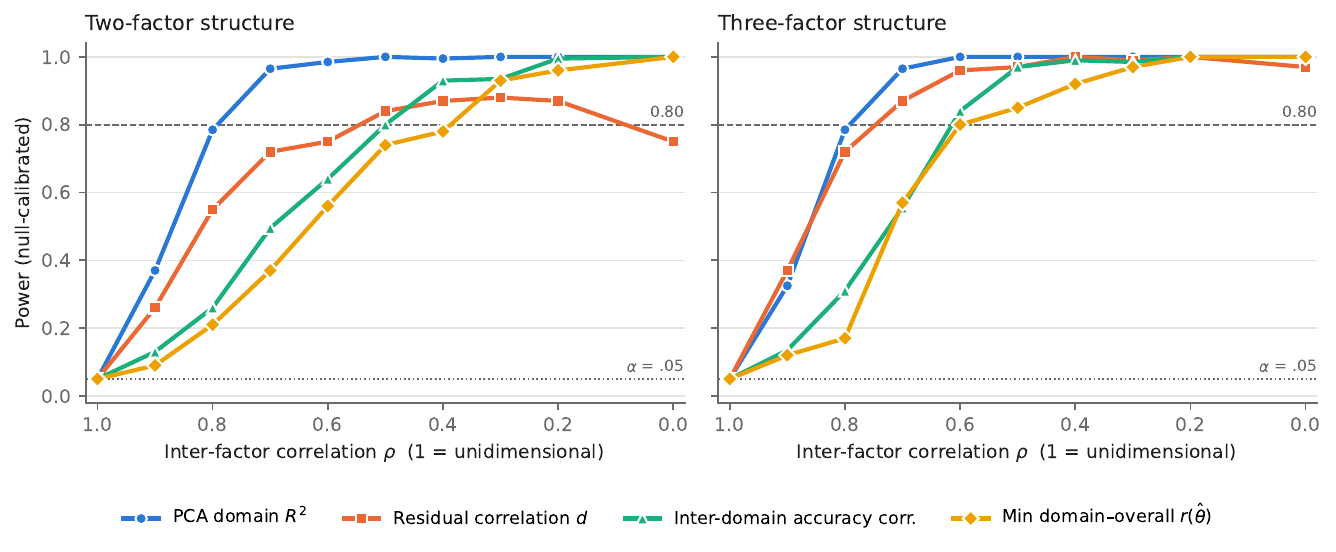}
\caption{Power of each statistic as domain abilities become more distinct ($\rho$: 1 to 0), for two-factor (left) and three-factor (right) structures. Dashed line: 0.80 power; dotted line: 5\% false-alarm rate.}
\label{fig:power}
\end{figure}

Figure~\ref{fig:power} shows the power of each dimensionality statistic for all simulated values of $\rho$, complementing the selected values in Table~\ref{tab:power}.

\end{document}